%% file: main.tex
\documentclass[10pt,twocolumn,letterpaper]{article}
\usepackage[dvipsnames,table,xcdraw]{xcolor}
\usepackage[accsupp]{axessibility}

\usepackage[pagenumbers]{cvpr} 

\input{preamble}

\definecolor{cvprblue}{rgb}{0.21,0.49,0.74}

\usepackage[pagebackref,breaklinks,colorlinks,citecolor=cvprblue]{hyperref}
\usepackage{graphicx}
\usepackage{epsfig}
\usepackage{amsmath}
\usepackage{amssymb}
\usepackage{booktabs}
\usepackage{graphicx}
\usepackage{multirow}
\usepackage{multicol}

\usepackage{tabularx}
\usepackage{adjustbox}
\usepackage{array}
\usepackage{multirow}
\usepackage{url}
\usepackage{booktabs}
\usepackage{array}
\usepackage{lipsum}
\usepackage{makecell} 
\usepackage{caption, subcaption}
\usepackage{bm}

\newcommand{\tableCellHeight}{1}
\newcommand{\tabstyle}[1]{
  \setlength{\tabcolsep}{#1}
  \renewcommand{\arraystretch}{\tableCellHeight}
  \centering
  \small
}

\newcommand{\tablestyle}[2]{\setlength{\tabcolsep}{#1}\renewcommand{\arraystretch}{#2}\centering\footnotesize}
\usepackage{float}
\definecolor{purple}{RGB}{230, 227, 254}
\definecolor{lightgreen}{RGB}{238, 252, 241}
\definecolor{lightred}{RGB}{231, 187, 187}
\definecolor{darkred}{RGB}{198, 129, 129}

\definecolor{tabhighlight}{HTML}{e5e5e5}

\def\ours{LLaMP}

\usepackage{pifont}
\newcommand{\cmark}{\ding{51}}%
\newcommand{\xmark}{\ding{55}}%
\def\eg{\emph{e.g.} }

\def\etal{\emph{et al.} }

\usepackage{enumitem}
\definecolor{lightgreen}{HTML}{E2F0D9}
\usepackage{amsmath}
\DeclareMathOperator*{\argmax}{argmax}
\newcolumntype{R}[2]{%
    >{\adjustbox{angle=#1,lap=\width-(#2)}\bgroup}%
    l%
    <{\egroup}%
}

\def\paperID{609} 
\def\confName{CVPR}
\def\confYear{2024}

\title{Large Language Models are Good Prompt Learners \\ for Low-Shot Image Classification}

\author{Zhaoheng Zheng \quad Jingmin Wei \quad Xuefeng Hu \quad Haidong Zhu \quad Ram Nevatia\\
Viterbi School of Engineering\\
University of Southern California\\
{\tt\small \{zhaoheng.zheng, jingminw, xuefengh, haidongz, nevatia\}@usc.edu}
}

\begin{document}
\maketitle
\input{sec/0_abstract}   
\input{sec/1_intro}
\input{sec/2_related-work}
\input{sec/3_approach}

\input{sec/4_exp}

\input{sec/5_con}

{
    \small
    \bibliographystyle{ieeenat_fullname}
    \bibliography{main}
}

\input{sec/X_suppl}

\end{document}

%% file: preamble.tex
\usepackage[dvipsnames]{xcolor}


%% file: sec/0_abstract.tex
\begin{abstract}

Low-shot image classification, where training images are limited or inaccessible, has benefited from recent progress on pre-trained vision-language (VL) models with strong generalizability, \eg CLIP. Prompt learning methods built with VL models generate text features from the class names that only have confined class-specific information. Large Language Models (LLMs), with their vast encyclopedic knowledge, emerge as the complement. Thus, in this paper, we discuss the integration of LLMs to enhance pre-trained VL models, specifically on low-shot classification. However, the domain gap between language and vision blocks the direct application of LLMs. Thus, we propose LLaMP, \textbf{L}arge \textbf{La}nguage \textbf{M}odels as \textbf{P}rompt learners, that produces adaptive prompts for the CLIP text encoder, establishing it as the connecting bridge. Experiments show that, compared with other state-of-the-art prompt learning methods, LLaMP yields better performance on both zero-shot generalization and few-shot image classification, over a spectrum of 11 datasets. Code will be made available at: \href{https://github.com/zhaohengz/LLaMP}{\texttt{https://github.com/zhaohengz/LLaMP}}.

\end{abstract}

%% file: sec/1_intro.tex
\section{Introduction}\label{sec:intro}
\begin{figure}[t]
    \centering    
    \begin{subfigure}[t]{\linewidth}
         \centering
         \includegraphics[width=\textwidth]{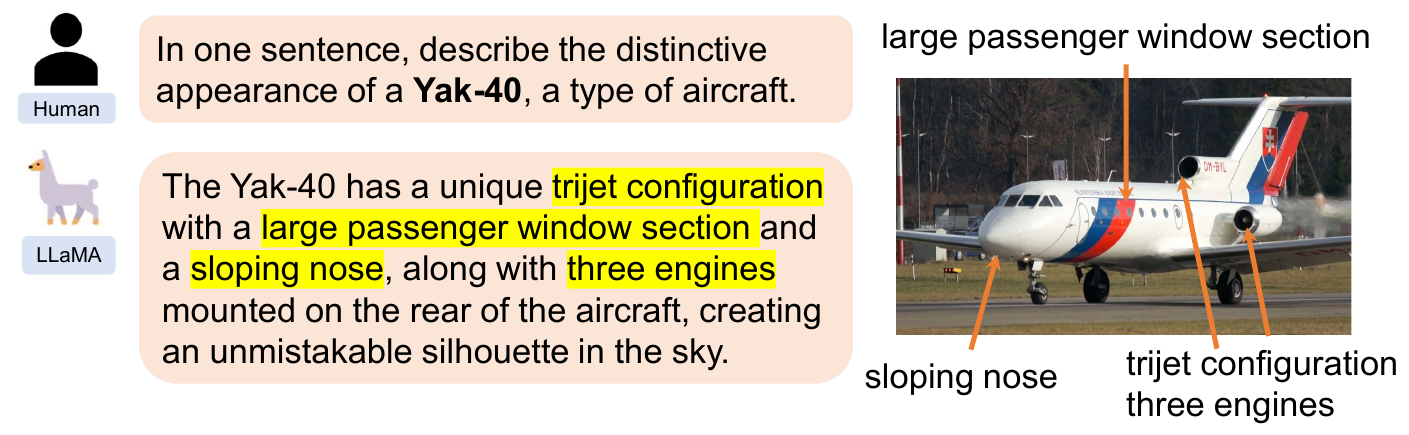}
         \caption{Visual Descriptions from an LLM.}
         \label{fig:teaser-demo}
     \end{subfigure}
    \begin{subfigure}[t]{\linewidth}
         \centering
         \includegraphics[width=\textwidth]{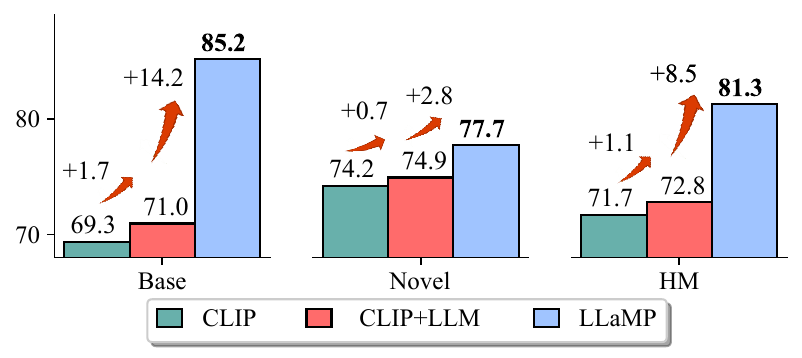}
         \caption{LLMs' Knowledge Boosts the Performance.}
         \label{fig:teaser-perf}
     \end{subfigure}    
     \caption{Demonstration of \ours: (a) LLMs can provide visual descriptions for fine-grained object categories; (b) Zero-shot base-to-novel generalization benefits from the LLM knowledge.}
     \vspace{-10pt}
\end{figure}
Low-shot image classification tasks, including few-shot and zero-shot variants, are to learn from a set of class names along with a limited or null set of images. Such capacities are crucial for the extension and generalization of vision systems. 
Vision-Language (VL) models trained on large-scale web data, such as CLIP \cite{radford2021learning} and ALIGN \cite{jia2021scaling}, provide a new paradigm due to their generalization capabilities that includes zero-shot classification, and have been used in recent work \cite{khattak2023maple,zhou2022conditional,zhou2022learning,khattak2023self,kan2023knowledge,lee2023read,lu2022prompt}.
Due to the scarcity of images for training, methods built for both tasks rely heavily on merely category names as the source of class-specific knowledge, resulting in a shortage of distinguishable descriptions.
Meanwhile, Large Language Models (LLMs), \eg GPT-4 \cite{openai2023gpt4} and LLaMA \cite{touvron2023llama,touvron2023llama2}, have demonstrated their encyclopedic knowledge and thus can provide linguistic visual descriptions for objects.
Here, we investigate how to leverage LLMs for low-shot image classification.

The emergence of prompt learning has provided an efficient way to adapt large pre-trained models. Previous work has explored various strategies to prompt vision-language (VL) models, including vision-conditioned text prompt learning \cite{zhou2022conditional}, joint VL prompt learning \cite{khattak2023maple} and self-regulated VL prompts\cite{khattak2023self}. On the text side, regardless of the learning strategy, learned prompt vectors are shared across all categories. The only difference among text inputs is the class name. In low-shot scenarios where visual data is limited, the extraction of class-specific knowledge from textual inputs becomes essential. However, the current paradigm, which relies on the CLIP text encoder to distinguish between class names, faces challenges, particularly with fine-grained target categories. For example, in FGVCAircraft \cite{maji2013fine}, the class name ``\texttt{Yak-40}'', can barely provide any information for recognizing the object. 

Large Language Models, trained with large text corpora, are good candidates to serve as the complement. As in Fig.~\ref{fig:teaser-demo}, being queried about ``\texttt{Yak-40}'', the LLM generates a sentence detailing the visual appearance of the Yak-40 that can be further parsed into noun phrases and integrated into text prompts, providing richer information, compared with the ordinary prompt. We also show in Fig.~\ref{fig:teaser-perf} that by simply incorporating noun phrases extracted from a LLM's response, the performance of the ordinary CLIP models is improved by more than 1\% without any training. Although recent prompt-learning based methods have shown notable improvements, it is non-trivial to apply them on textual visual descriptions generated by LLMs. Thus, instead of directly taking LLM generations as the textual input, we aim at producing class-specific representations by adapting LLMs to low-shot image classification.

One challenge of the adaption is the domain gap between vision and language. When trained exclusively with textual corpora, the latent feature space of a LLM significantly diverges from that of its visual counterpart. Even worse, the data scarcity under the low-shot scenario make it virtually impossible to align two spaces through plain contrastive loss. We argue that, the CLIP text encoder, which is trained to project features from the language domain into the joint VL domain, can serve as the bridge. Thus, we propose the \textbf{LLaMP} framework, \textbf{L}arge \textbf{La}nguage \textbf{M}odels as \textbf{P}rompt learners, which leverages LLMs to learn informative prompts for CLIP models. In \ours, we treat LLMs as the prompt learner of the CLIP text encoder. More specifically, for each object category, \ours{} extracts corresponding knowledge from the LLM and yields class-specific prompt vectors, which are further combined with class-agnostic prompt embeddings (as in previous approaches), and encoded by the CLIP text encoder. We design an efficient tuning pipeline to avoid fully fine-tuning the language model while performing effective adaptation.

Following the protocol in \cite{zhou2022conditional,zhou2022learning}, we evaluate \ours{} with two typical scenarios: zero-shot base-to-novel generalization \cite{zhou2022learning} and few-shot image classification. For each scenario, we run \ours{} with 11 datasets covering a spectrum of tasks. On average, \ours{} achieves a 1.3\% boost on the harmonic mean against the state-of-the-art PSRC \cite{khattak2023self}, and 9.6\% over the ordinary CLIP \cite{radford2021learning}, on base-to-novel generalization. We also observe an average improvement of 0.94\% on 16-shot image classification.

In summary, our approach makes use of Large Language Models to improve performance in low-shot image classification scenarios. The main contributions are: i) To the best of our knowledge, we are the first to investigate how to use the encyclopedic knowledge inherent in Large Language Models (LLMs) to enhance low-shot image classification; ii) We design a framework, \ours, to effectively adapt LLMs for image classification, without training the entire language model, and achieve state-of-the-art in both few-shot and zero-shot settings; iii) We conduct extensive analysis investigating the effectiveness of each components of \ours, and discuss the optimal setup for LLM-aided image classification. 

%% file: sec/2_related-work.tex
\section{Related Work}

\textbf{Large Language Models (LLMs).} Recent years have witnessed remarkable progress in scaling up the size and capabilities of LLMs. Zhang \etal \cite{zhang2022opt} first introduced a suite of transformers pre-trained at scale, followed by PaLM \cite{chowdhery2022palm}. ChatGPT/GPT-4 \cite{chatgpt, openai2023gpt4} emerged as a milestone conversational model, demonstrated impressive conversational abilities as a generalist model. Vicuna \cite{vicuna2023} further advanced by learning from ChatGPT, while LLaMA \cite{touvron2023llama} demonstrates that larger scale training yields stronger foundation models. The subsequent LLaMA-2 \cite{touvron2023llama2} and PaLM-2 \cite{anil2023palm} achieved further gains in scale, efficiency and reasoning. Most recently, Almazrouei \etal \cite{almazrouei2023falcon} released Falcon, a 400B model.

\textbf{Zero-Shot Learning (ZSL).} ZSL stands in contrast to traditional fully-supervised paradigms. Instead of relying on direct visual training samples, it leverages side information that can be drawn from a multitude of non-visual domains, including attributes \cite{lampert2013attribute}, word embeddings \cite{wang2018zero,socher2013zero}, and descriptive texts \cite{reed2016learning}. Zhang \etal \cite{zhang2017learning} designed an embedding model to bridge the gap between seen and unseen categories. Concurrently, studies like \cite{chen2021free,xian2018feature,zhu2018generative} have spotlighted that generative models can produce features for unseen categories. Moreover, Graph Convolution Networks (GCN) \cite{KipfW17} has been explored in research such as \cite{wang2018zero,kampffmeyer2019rethinking} for further generalization. 

\textbf{Prompt Learning.} With the progress in large-scale vision-language models, such as CLIP \cite{radford2021learning} and ALIGN \cite{jia2021scaling}, which reveal their capacity in zero-shot transferability, prompt learning has emerged as an efficient learning scheme, where learnable prompts are appended to the input to fine-tune models. For low-shot image classification, CoOp \cite{zhou2022learning} and CoCoOp \cite{zhou2022conditional}, which modeled context words as learnable vectors to automate prompt engineering, have shown significant improvements over regular CLIP. MaPLe \cite{khattak2023maple} further employed a hierarchical multi-modal prompting strategy across transformer blocks for progressive feature modeling. Kan \etal \cite{kan2023knowledge} incorporated external knowledge by designing knowledge-aware prompts and adaptation head for better generalization. Lee \etal \cite{lee2023read} used masked attention to prevent internal representation shift for better generalization. Khattak \etal \cite{khattak2023self} further improved prompt learning by guiding prompts to balance task-specific and task-agnostic knowledge via mutual agreement maximization and prompt ensemble. 

%% file: sec/3_approach.tex
\section{Approach}
\begin{figure*}
    \centering
    \includegraphics[width=\textwidth]{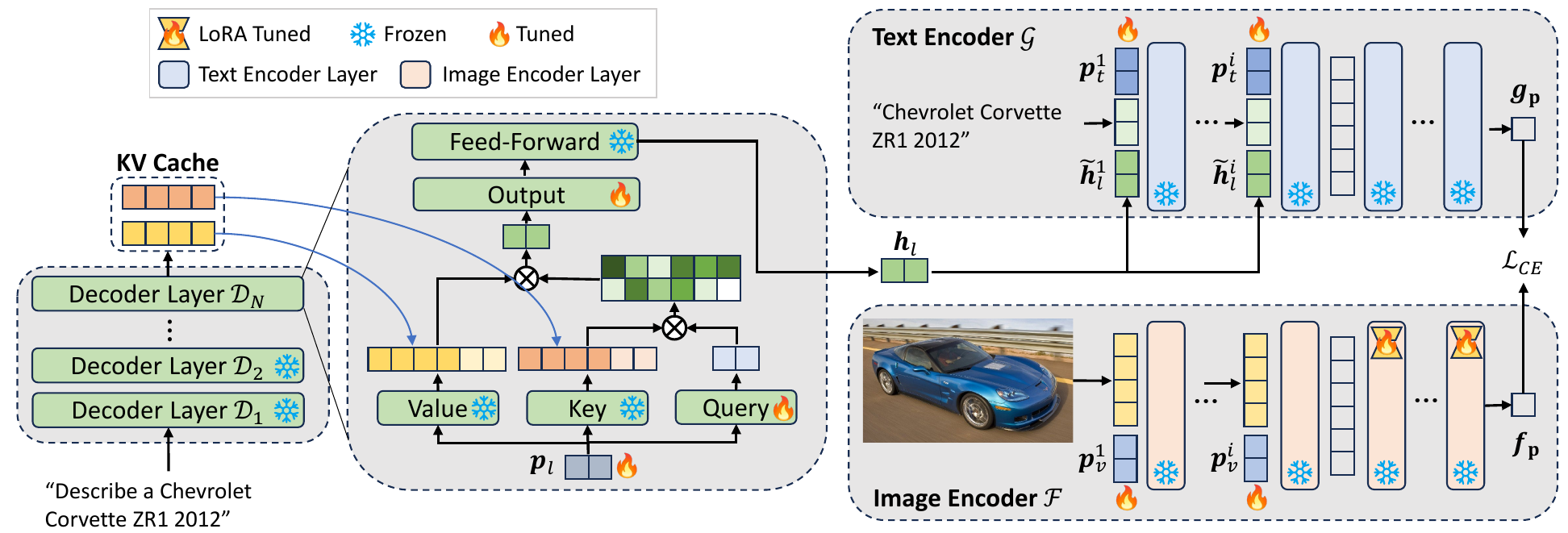}
    \caption{An Overview of the \ours{} Framework: We first generate the knowledge cache by passing the query prompt through the LLM $\mathcal{D}$ and use the knowledge cache to encode $\bm{p}_l$, resulting the adaptive prompts $\bm{\Tilde{h}}^i_l=W\bm{h}^i_l + b^i$ for the CLIP text encoder. $\bm{\Tilde{h}}_l$ is combined with regular learnable prompts of $\mathcal{G}$ to generate the final text feature vector $\bm{g_p}$. The image feature vector $\bm{f_p}$ is obtained through a hybrid-tuning strategy combining prompt learning and low-rank adaptation (LoRA).}
    \vspace{-15pt}
    \label{fig:main}
\end{figure*}

\subsection{Preliminaries}

Similar to previous CLIP-based learning approaches, we consider the classification problem as an image-text matching problem. We denote the image encoder and the text encoder, in CLIP-like models, as $\mathcal{F}$ and $\mathcal{G}$, parameterized by $\theta_{\mathcal{F}}$ and $\theta_{\mathcal{G}}$, respectively. An input image $\bm{x} \in \mathbb{R}^{C\times H \times W}$ is split into $M$ equal-sized patches which are converted into a sequence of embeddings $\bm{\Tilde{x}} = \{\bm{e}_{cls}, \bm{e}_{1}, \bm{e}_{2}, \dots, \bm{e}_M$\}. The visual input sequence $\bm{\Tilde{x}}$ is encoded by the image encoder, producing the image feature $\bm{f} = \mathcal{F}(\bm{\Tilde{x}})$. On the text side, the text label $y$ and the associated name is formatted as ``\texttt{A photo of [STH]}'' and tokenized into a sequence of tokens $\bm{\Tilde{y}} = \{\bm{t}_{bos}, \bm{t}_{1}, \bm{t}_{2}, \dots, \bm{t}_{L}, \bm{t}_{eos}\}$, where $L$ is the length of input tokens. The input sequence is then encoded into $\bm{g} = \mathcal{G}(\bm{\Tilde{y}})$. For image classification, target class labels $\{1, 2, \dots, C\}$ are encoded into text features $g_i$. The classification is done by picking the class that has the highest similarity with the vision feature: $\hat{y} = \argmax_i \, \mathcal{C}(\bm{f}, \bm{g}_i)$, where $\mathcal{C}$ is the softmax cosine-similarity function $\mathcal{C}(\bm{f},\bm{g}) = \frac{\exp{(\bm{f} \cdot \bm{g} / \tau)}}{\sum_{j=1}^{C}\exp{(\bm{f} \cdot \bm{g}_j / \tau)}}$ with temperature $\tau$.

\textbf{Multimodal Prompting Learning. } Given the size of the CLIP model, fine-tuning the entire model becomes infeasible. As both image and text encoders are built with standard transformer architecture, prompt learning, which tunes the model by combining trainable prompts with hidden states has been applied on the text encoder \cite{zhou2022conditional,zhou2022learning}, the image encoder \cite{jia2022visual,wang2022dualprompt,wang2022learning}, or both \cite{khattak2023maple,khattak2023self,rasheed2023fine}. Similar to \cite{khattak2023self,rasheed2023fine}, we build our method following the vision-language prompting paradigm, with deep prompting \cite{jia2022visual,khattak2023self}, which not only insert prompts to the input layer, but to later encoder layers.

More specifically, for each transformer layer that takes prompts, we define $V$ learnable visual prompts $\bm{p}_v=\{p_{v}^1, p_{v}^2, \dots, p_{v}^{V}\}$ and $T$ learnable language prompts $\bm{p}_{t} =\{p_{t}^1, p_{t}^2, \dots, p_{t}^{T}\}$. For the $i$-th vision encoder layer, visual prompts $\bm{p}_v^{i}$ are appended to input embeddings: $\bm{\Tilde{x}_p}^{i}=\{\bm{e}_{cls}^i, \bm{e}_{1}^i, \bm{e}_{2}^i, \dots, \bm{e}_M^i, \bm{p}_v^i\}$. The prompt-augmented vision feature, $\bm{f_p}=\mathcal{F}(\bm{\Tilde{x}_p})$, is produced by jointly encoding prompts and the image.
As the ViT \cite{dosovitskiy2021an} architecture in CLIP adopts the bi-directional attention mechanism, the placement of $\bm{p}_v$ has no effect on $\bm{f_p}$. On the language side, prompts are concatenated with the input of the $i$-th text encoder: $\bm{\Tilde{y}_p}^{i}=\{\bm{t}_{bos}^i, \bm{p}^i_t, \bm{t}_{1}^i, \bm{t}_{2}^i, \dots, \bm{t}_{L}^i,  \bm{t}_{eos}^i\}$. $\bm{\Tilde{y_p}}$ is further processed by the text encoder, resulting in the prompt-augmented language feature $\bm{g_p}=\mathcal{G}(\bm{\Tilde{y}_p})$. More specifically, prompts to the first layer $\bm{p}_t^1$ are initialized with the embeddings of ``A photo of a''. 

\textbf{Low-Rank Adaptation \cite{hu2021lora} (LoRA).} As a parameter-efficient tuning technique, LoRA is designed to adapt large transformer model without updating original model weights. The LoRA technique is, in particular, applied to linear projection layers. More specifically, for a linear layer with weight $W_0 \in \mathbb{R}^{d \times k}$, LoRA creates $\Delta W$ by learning two low rank matrices $B \in \mathbb{R}^{d \times r}$ and $A \in \mathbb{R}^{r \times k}$:
\begin{equation}
\bm{h} = (W_0 + \Delta W)\bm{x}  = W_0 \bm{x} + BA\bm{x}.
\end{equation}
We adopt a hybrid tuning scheme on the vision encoder, which performs prompt learning on the first few layers and applies LoRA on the rest. 

\subsection{Adaptive Prompt Learning with LLMs}

The goal of prompt tuning is to find a set of optimal prompts $\bm{p}=\{\bm{p}_v, \bm{p}_t\}$ which maximizes the log likelihood of $P(x,y|\theta_\mathcal{F},\theta_\mathcal{G})$ over target downstream distribution  $(\bm{x},\bm{y}) \sim (\bm{X}, \bm{Y})$:
\begin{equation}
     \bm{p} = \argmax_{\bm{p}}  \mathbb{E}_{(\bm{x},\bm{y})\sim (\bm{X}, \bm{Y})} \log\mathcal{C}(\mathcal{F}(\bm{x};\bm{p}_v), \mathcal{G}(\bm{y};\bm{p}_t))
     \label{eqn:prompt}
\end{equation}

However, the $\bm{p}$ optimized through Eqn.~\ref{eqn:prompt} has two issues. First, $\bm{p}$ is shared for all categories for the downstream task, while the optimal prompt for each category might be different. Second, in low-shot scenarios, $\bm{p}$ are usually empirically estimated from a limited training-set $\bm{X}^{train}$ with limited categories $\{1,2,...,C^{base}\}$, and therefore such $\bm{p}$ can often be over-fitted to the small training-set $\bm{X}^{train}$ and fail to generalize to novel categories outside $\{1,2,...,C^{base}\}$. 

To overcome these problems, we propose to learn a meta function on the language side $\bm{p}_t=\Theta(y)$ which can adaptively estimate the optimal prompt for each category. 
An intuitive way to estimate proper prompts $\bm{p}$ for category name $y$ is to take advantage of the knowledge of the pre-trained Large Language Models (LLM) $\mathcal{D}$ and extract discriminative descriptions of category $y$. For example, given the input text $\bm{z}$:``Describe \{y\}'', 
\begin{equation}
    \bm{p}_t = \{p_1, p_2, ..., p_k\} = \mathcal{D}(\bm{z}).
\end{equation}
while $p_i$ being sequentially generated by $\mathcal{D}$ such that 
\begin{equation}
\begin{aligned}
    p_i &= \mathcal{D}(\bm{z}, t_1,...,t_{i-1}) = \mathcal{D}^{(i)}(
    \bm{z}) \\
    t_i & = \mathcal{M}(p_i),
\end{aligned}
\end{equation}
where $\mathcal{D}^{(i)}$ is the $i$-th forward iteration of $\mathcal{D}$, and $\mathcal{M}$ maps continuous hidden states into discrete language tokens. To accelerate the process and to obtain $p$ in one pass, we approximate the above process with $K$ learnable prompts $\bm{p}_l = \{\theta_1,...,\theta_K\}$ so that 
\begin{align}
    \bm{p}_t &= \Theta(y) = \mathcal{D}(\{\theta_1,...,\theta_K\}|\bm{z})
\end{align}

\textbf{Discussion.} While Large Language Models (LLMs) possess robust foundational knowledge within the linguistic domain, it is not feasible to directly substitute the text encoder of CLIP with an LLM. The reason lies in the inherent divergence between the LLM's latent space, which is purely language-oriented, and the image-focused latent space of vision encoders. Attempting a direct alignment via contrastive learning would require an extensive dataset that is typically beyond the scope of low-shot learning. To bridge this gap, we introduce \ours—an adaptive prompt learning framework that leverages the LLM to craft class-specific prompt vectors, to reinforce the text encoder for low-shot image classification.

\subsection{The LLaMP Framework}
\label{sec:llamp}

Fig.~\ref{fig:main} shows an overview of the \ours{} framework. For convenience, we denote the decoder-only LLM as $\mathcal{D}$. The input to the decoder $\mathcal{D}$ consists of two components: textual prompts $\bm{y}$ in the form of sentences, tokenized as $\bm{\Tilde{y}}$, and learnable prompts $\bm{p}_l$. We append prompt embeddings to the end of the input sequence and obtain the last hidden states of $\mathcal{D}$ as the feature $\bm{h}_l$:
\begin{equation}
\bm{h}_l = \mathcal{D}(\bm{\Tilde{y}}, \bm{p}_l)[L+1:L+K], L = \text{Length}(\bm{\Tilde{y}}).
\end{equation}
Hidden states of $\mathcal{D}$ are then mapped to the input space of the CLIP text encoder by the projection matrix $W \in \mathbb{R}^{d_1 \times d_2}$, where $d_1$ and $d_2$ are respectively the hidden sizes of the LLM $\mathcal{D}$ and the CLIP text encoder $\mathcal{G}$. A set of prompt-specific biases $b \in \mathbb{R}^{K \times d_2}$ are added: 
\begin{equation}
    \bm{\Tilde{h}}_l = W \bm{h}_l + b
\end{equation}

We combine $\bm{\Tilde{h}_l}$ from LLM with regular learnable prompts, as in previous approaches \cite{khattak2023self}, to construct the input for CLIP text encoder. Similar to deep prompting \cite{jia2022visual,khattak2023self}, we create layer-specific prompts through different $W$ matrices and $b$ vectors. For the $i$-th layer, we let $\bm{\Tilde{h}}_l^i = W^i \bm{h}_l + b^i$ and the entire sequence is constructed as 
\begin{equation}   
 \bm{\Tilde{y}_{l}}^{i}=\{\bm{t}_{bos}^i, \bm{p}^i_t, \bm{t}_{1}^i, \bm{t}_{2}^i, \dots, \bm{t}_{L}^i,  \bm{\Tilde{h}_l^i}, \bm{t}_{eos}^i\}
\end{equation}

\textbf{LLM Knowledge Cache.}
A Large Language Model (LLM), as implied by its name, typically comprises billions of parameters. For example, the most compact LLaMA \cite{touvron2023llama,touvron2023llama2} model has 7B parameters. Thus, even performing prompt learning on a LLM become impractical. The memory consumption to store gradients for back propagation can go beyond the limit of mainstream GPUs. Instead, the causal attention mechanism inherent in decoder-only LLMs, where the embedding of an input token only depends on the preceding tokens, facilitates a feasible workaround. 

As previously mentioned, the prompt embeddings $\bm{p}_l$ are appended to the end of text tokens $\bm{\Tilde{y}}$. According to the causal attention mechanism, $\bm{\Tilde{y}}$ is encoded independently of $\bm{p}_l$. Thus, we design a two-stage process, where we create the LLM knowledge cache by passing $\bm{\Tilde{y}}$ through $\mathcal{D}$ and leverage the cache to convert $\bm{p}_l$ into class-specific embeddings for the CLIP text encoder $\mathcal{G}$.

To compute the attention of a token, the only dependency is the \textit{Key} and \textit{Value} vectors from the preceding tokens. Thus, we adopt the KV-cache \cite{pope2023efficiently,wolf-etal-2020-transformers}, a technique used in inference acceleration of LLMs, to create the knowledge cache. At the first stage, we pass text tokens $\bm{\Tilde{y}}$ through the language model $\mathcal{D}$ and save the \textit{Keys} and \textit{Values} as the knowledge cache for the second stage. Once computed, the knowledge cache remains fixed throughout the entire training process and bears the information that is needed for further computation. Thus, in LLaMP, we leverage the knowledge cache obtained at the first stage to generate class-specific prompt embeddings.

At the second stage, we create class-specific prompt embeddings from the pre-computed knowledge cache. As $\bm{p}_l$ is not initialized in the natural language domain, it need not pass through the entire LLM; instead, we insert those prompts $\bm{p}_l$ to the last layer of the LLM $\mathcal{D}_{N}$. It is achieved by encoding them alongside the cache from $\bm{\Tilde{y}}$, as in
\begin{equation}
    \bm{H}_l = \mathcal{D}_{N}(\bm{K}_{\bm{\Tilde{y}}}, \bm{V}_{\bm{\Tilde{y}}}, \bm{p}_l),
    \label{eqn:last}
\end{equation}
where $\bm{K}_{\bm{\Tilde{y}}}, \bm{V}_{\bm{\Tilde{y}}}$ represent the knowledge cache. This design enables \ours{} to efficiently learn informative prompt embeddings for the CLIP encoder $\mathcal{G}$. It accomplishes this by incurring modest training costs, compared with training  the entire LLM. Simultaneously, it maintains the essential knowledge inherent in the LLM decoder $\mathcal{D}$.

\textbf{Training Targets of LLaMP.} Although the training strategy in Eqn.~\ref{eqn:last} has reduced the number of learnable parameters, a full decoder inside a LLM still consists of an enormous number of parameters. For example, one layer in LLaMA-7B bears 200M parameters, making training of the entire layer costly. As the goal is to leverage the knowledge from LLM, altering a full layer can lead to the loss of knowledge. Thus, as shown in Fig.~\ref{fig:main}, a typical decoder layer has two major components: the self attention module, consisting \textit{Query}, \textit{Key}, \textit{Value} and \textit{Output} projection layers, and the Feed-Forward Network (FFN). \ours{} targets the \textit{Query} and \textit{Output} projection layer inside the self-attention module. By updating the \textit{Query} layer, LLM prompts $\bm{p}_l$ are learned to distill pertinent information from the knowledge  cache and the \textit{Output} layer projects it to the latent space. We keep the \textit{Key} and \textit{Value} layers frozen to ensure the alignment between $\bm{p}_l$ and knowledge cache. We leave the FFN unchanged to preserve the knowledge. Further discussions regarding these choices are made in Sec.~\ref{exp:ablation}.

\textbf{Textual Priors from Pre-Generated Responses.} We extend the initial prompt, ``\texttt{In one sentence, describe the distinctive appearance of [STH]}'', by incorporating the response generated by the language model into the input sequence. This approach enriches the base content: the generated text provides a clear and explicit description of the object’s appearance, acting as a valuable informative prior for language model adaptation. However, it is common for responses from an LLM to include filler words like ``sure'' for sentence structure coherence. To refine the input, we parse the noun phrases from the LLM’s response through spaCy \cite{Honnibal_spaCy_Industrial-strength_Natural_2020}, an NLP engine, and merge them with the initial prompt, forming a more focused and informative language prior.

\textbf{Textual Augmentations.} Following the insights of Khattak et al. \cite{khattak2023self}, which highlight the performance benefits of diverse textual inputs, we aim to further augment the text inputs used in the CLIP text encoder. Our approach, building upon the methods in \cite{khattak2023self, zhou2022conditional}, incorporates hand-crafted templates and expands their diversity through a two-step process: i) We introduce noun phrases into the existing templates for CLIP, for example, transforming ``\texttt{A photo of [STH]}'' to ``\texttt{A photo of [STH] with [NP]}'', thereby enriching the descriptive content; ii) We create a variety of new prompt templates for the LLM similar to ``\texttt{In one sentence, describe the distinctive appearance of [STH]}'' through GPT-4 \cite{openai2023gpt4}, to further diversify the text input.

\subsection{Training and Inference}
Similar to PSRC \cite{khattak2023self}, our objective function consists of three components: The main cross-entropy loss $\mathcal{L}_{CE}$, feature-level L1 regularization $\mathcal{L}_{l1}$, and soft distillation loss $\mathcal{L}_{dist}$. Given $\mathcal{C}$ training categories and $\mathcal{N}$ training samples, $\mathcal{L}_{CE}$ is defined as 
\begin{equation}
\mathcal{L}_{CE} = -\frac{1}{\mathcal{N}}\sum_{i} \log \frac{\exp{(\bm{f_p}^i \cdot \bm{g_p}^i / \tau)}}{\sum\exp{(\bm{f_p}^i \cdot \bm{g_p}^j / \tau)}}.
\end{equation}
The L1 regularization is computed between learned features $\bm{f_p}, \bm{g_p}$ and pre-trained CLIP features $\bm{\hat{f}}, \bm{\hat{g}}$: 
\begin{equation}
    \mathcal{L}_{l1} = \frac{1}{\mathcal{N}}\sum_{i}\lambda_v|\bm{f_p}^i - \bm{\hat{f}}^i| + \frac{1}{C}\sum_i\lambda_t |\bm{g_p}^i - \bm{\hat{g}}^i|,
\end{equation}
where $\lambda_v$ and $\lambda_t$ are coefficients. The prediction of \ours{} is further bound by the KL-Divergence between predicted distributions of \ours{} and vanilla CLIP:
\begin{equation}
\mathcal{L}_{dist} = \lambda_{dist} D_{KL}(\bm{f_p} \cdot \bm{g_p}, \bm{\hat{f}} \cdot \bm{\hat{g}}).
\end{equation}
We sum all three losses up as the final objecttive function: $\mathcal{L} = \mathcal{L}_{CE} + \mathcal{L}_{l1} + \mathcal{L}_{dist}.$

During training, we randomly sample one LLM template as the input of \ours{} for each batch. For inference, we compute the probability distribution predicted from each input template and average them.

%% file: sec/4_exp.tex
\section{Experiments}
\input{sec/main_table_compact}
\subsection{Experiment Setup}

\textbf{Datasets.} Similar to previous work \cite{khattak2023self,khattak2023maple,zhou2022conditional}, in our study, we evaluate \ours{} performance over a spectrum of classification tasks with 11 datasets, including ImageNet \cite{deng2009imagenet} and Caltech101 \cite{fei2004learning} for generic image classification, OxfordPets \cite{parkhi2012cats}, StanfordCars \cite{krause20133d}, Flowers102 \cite{nilsback2008automated}, Food101 \cite{bossard2014food}, and FGVCAircraft \cite{maji2013fine} for fine-grained classification, SUN397 \cite{xiao2016sun} for scene recognition, UCF101 \cite{soomro2012ucf101} for action recognition, DTD \cite{cimpoi2014describing} for texture classification, and EuroSAT \cite{helber2019eurosat} for satellite image recognition. 

\begin{table*}[!t]
    \tabstyle{2.5pt}
        \begin{tabular}{l c ccccccccccc}
        \toprule
        & \multicolumn{12}{c}{16-Shot Classification} \\ \cmidrule(lr){2-13}
        & \footnotesize\rotatebox{60}{\emph{Average}} & \footnotesize\rotatebox{60}{ImageNet \cite{deng2009imagenet}} & \footnotesize\rotatebox{60}{Caltech \cite{fei2004learning}} & \footnotesize\rotatebox{60}{Pets \cite{parkhi2012cats}} & \footnotesize\rotatebox{60}{Cars \cite{krause20133d}} & \footnotesize\rotatebox{60}{Flowers \cite{nilsback2008automated}} & \footnotesize\rotatebox{60}{Food \cite{bossard2014food}} & \footnotesize\rotatebox{60}{Aircraft \cite{maji2013fine}} & \footnotesize\rotatebox{60}{SUN397 \cite{xiao2016sun}} & \footnotesize\rotatebox{60}{DTD \cite{cimpoi2014describing}} & \footnotesize\rotatebox{60}{EuroSAT \cite{helber2019eurosat}} & \footnotesize\rotatebox{60}{UCF101 \cite{soomro2012ucf101}}  \\
        \midrule
        CLIP \cite{radford2021learning}  & 78.79 (65.02)  & 67.31 & 95.43 & 85.34 & 80.44 & 97.37 & 82.90 & 45.36 & 73.28 & 69.96 & 87.21 & 82.11 \\
        CoOp \cite{zhou2022learning} & 79.89 (73.82) & 71.87 & 95.57 & 91.87 & 83.07 & 97.07 & 84.20 & 43.40 & 74.67 & 69.87 & 84.93 & 82.23 \\
        CoCoOp \cite{zhou2022conditional} & 74.90 (70.70) & 70.83 & 95.16 & 93.34 & 71.57 & 87.84 & 87.25 & 31.21 & 72.15 & 63.04 & 73.32 & 78.14 \\
        MaPLe \cite{khattak2023maple} & 81.79 (75.58) & 72.33 & 96.00 & 92.83 & 83.57 & 97.00 & 85.33 & 48.40 & 75.53 & 71.33 & 92.33 & 85.03  \\
        PSRC \cite{khattak2023self} & 82.87 (77.90) & 73.17 & 96.07 & 93.67 & 83.83 & 97.60 & 87.50 & 50.83 & \textbf{77.23} & 72.73 & \textbf{92.43} & 86.47  \\
        \midrule
        \rowcolor{tabhighlight}
        \ours  & \textbf{83.81} (\textbf{78.50}) & \textbf{73.49} & \textbf{97.08} & \textbf{94.21} & \textbf{86.07} & \textbf{98.06} & \textbf{87.62} & \textbf{56.07} & 77.02 & \textbf{74.17} & 91.31 & \textbf{86.84} \\
    
        \bottomrule
        \end{tabular}
            \caption{Few shot classification results with 16 shots. Numbers in the bracket indicate the average performance over 1/2/4/8/16 shots.}
        \label{tab:xd}
        \vspace{-5mm}
\end{table*}

\textbf{Scenarios \& Metrics.} We evaluate \ours{} on two typical low-shot scenarios: zero-shot base-to-novel generalization and few-shot image classification. In zero-shot base-to-novel generalization, the base classes are seen during training, while the novel classes are unseen. We measure models performance through accuracies of base and novel classes, and the harmonic mean of the two. For few-shot classification, we assess the accuracy with 16 shots per class.

\textbf{Implementation Details.} We build \ours{} through the PyTorch \cite{paszke2019pytorch} framework. All models are trained with 2 NVIDIA A100 40GB GPUs. For \ours, we adopt LLaMA2-7B \cite{touvron2023llama2} as the language model $\mathcal{D}$, and ViT-B/16 \cite{dosovitskiy2021an} as the image encoder, following \cite{khattak2023self,khattak2023maple,zhou2022conditional,zhou2022learning}. On the text side, we set prompt learning depth to 9. To tune the vision encoder, we adopt the hybrid tuning scheme which performs deep prompt learning on the first 6 layers and LoRA on the rest. Similar to \cite{hu2021lora}, LoRA is applied to the \textit{Query} and \textit{Value} projection layers inside attention modules. The number of $\bm{p}_l$ prompts, $K$, is set to 16. We set a global learning rate of $2\text{E-}4$ with a batch size of 8. The learning rate of LoRA modules is set to $2\text{E-}5$. $\lambda_t, \lambda_v$ and $\lambda_{dist}$ are set to 25, 10 and 2.5, respectively. 

\subsection{Quantitative Evaluation}

\textbf{Zero-Shot Base-to-Novel Generalization.} \ours{} outperforms existing state-of-the-art prompt learning methods on most metrics of 11 classification datasets in the base-to-novel generalization benchmark. As shown in Tab.~\ref{table:comparision_with_cocoop}, compared to the latest model PSRC \cite{khattak2023self}, \ours{} achieves average gains of 0.90\% in base accuracy, 1.61\% in novel accuracy, and 1.30\% in harmonic mean on average. Moreover, \ours{} consistently achieves higher harmonic means (HM) compared to other models. These improvements indicate that our approach better balances performance on base and novel data, thus achieving stronger generalization compared to the prior prompt learning techniques.

In particular, \ours{} excels in fine-grained datasets requiring detailed analysis. On \textit{FGVCAircraft}, \ours{} surpasses PSRC by 4.57\% on base accuracy and 1.75\% on HM, highlighting its strong understanding of detailed aircraft features. Furthermore, on \textit{EuroSAT}, \ours{} achieves improvements of 9.76\% and 5.28\% on novel accuracy and HM, respectively. We also observe similar performance gains on \textit{StanfordCars}, where \ours{} outperformns by 3.29\% on base accuracy and 1.31\% on HM. The information embedded in LLM enables \ours{} to capture and utilize the rich semantic information necessary for distinguishing between closely related categories.

\textbf{Few-Shot Classification.} \ours{} also achieves improvements across these classification datasets in few-shot classification tasks. As in Tab.\ref{tab:xd}, with an average classification accuracy of 83.81\%. Notably, on \textit{FGVCAircraft} and \textit{StanfordCars},  \ours{} shows a significant improvement over PSRC, further demonstrating that the knowledge from language models benefits the recognition of fine-grained object categories, which aligns with our observation on zero-shot base-to-novel generalization. Moreover, on \texttt{DTD}, where MaPLe and PSRC achieve around 72\% accuracy, \ours{} achieves a higher accuracy of 74.17\%, underscoring its ability to recognize textures.

\subsection{Ablation Study}

\label{exp:ablation}

\begin{table}[t]
    \centering
    \begin{tabular}{lc|cc|c}
        \toprule
        Method & LLM & Base & Novel & HM \\
        \midrule
        \multirow{2}{*}{CLIP}  & & 69.34 & 74.22 & 71.70 \\
        & \checkmark & 70.95 & 74.93 & 72.79\\
        \midrule
        \multirow{2}{*}{\ours}  & & 82.21 & 76.44 & 79.22 \\
        & \checkmark &\cellcolor{tabhighlight}\textbf{85.16} &\cellcolor{tabhighlight}\textbf{77.71} & \cellcolor{tabhighlight}\textbf{81.27} \\
    \bottomrule
    \end{tabular}
    \caption{Ablation study on the LLM Knowledge.}
    \vspace{-5pt}
    \label{tab:abl-llm}
\end{table}

\begin{table}[t]
        \centering
        \resizebox{\linewidth}{!}{
        \begin{tabular}{cccc|c|cc|c}
            \toprule
            LP & QO & KV & FFN & \% &Base & Novel & HM \\
            \midrule
            \cmark & & & & .03 & 85.00 & 77.29 & 80.96 \\
            \cmark & \cmark & \cmark  &  & 33 & 85.20 & 77.45 & 81.14 \\
            \cmark & \cmark & & \cmark & 83 & 85.05 & \textbf{77.73} & 81.22\\
            \cmark & \cmark & \cmark & \cmark & 100 &\textbf{85.23} & 77.56 & 81.21 \\
            \midrule
            \rowcolor{tabhighlight}
            \cmark  & \cmark & & & 17& 85.16 &77.71 & \textbf{81.27} \\
        \bottomrule
        \end{tabular}
        }
        \caption{Ablation study on the Training Strategy. ``\%'' indicates the ratio of parameters trained compared to fully tuning a layer.}
        \vspace{-10pt}
        \label{tab:abl-training}
    \end{table}

\textbf{Is the knowledge from LLM helping?} In Tab.~\ref{tab:abl-llm}, we show that the knowledge from LLM benefits in both ways: Without training, performance of ordinary CLIP model can be improved by introducing noun phrases; The \ours{} framework shows further improvement after training.

Noun phrases are parsed from the LLM's responses the prompt of ``\texttt{Describe [STH]}''. We then use the template, ``\texttt{A photo of [STH] with [NP]}'' to generate the NP-augmented text embedding for CLIP. We take the average of all augmented embeddings for classification. In Tab.~\ref{tab:abl-llm} we show that even ordinary CLIP can benefit from incorporating LLMs' knowledge.

Furthermore, the comparison between \ours{} and \ours{} without the LLM indicates that merely integrating LoRA \cite{hu2021lora} to the vision encoder is not beneficial. The ``\ours{} without LLM'' is essentially an ordinary prompting learning model plus LoRAs in the vision encoder. We show that the improved vision encoding capacity only benefits when the quality of text embeddings s are enhanced by incorporating LLMs' knowledge through \ours.

\textbf{Decoder Training strategy.} We categorize trainable parameters of $\mathcal{D}_{N}$ into four groups: learnable prompts (LP), \textit{Query} and \textit{Output} projections (QO), \textit{Key} and \textit{Value} projections (KV), and the feed-forward network (FFN). Tab.~\ref{tab:abl-training} indicates \ours{} can achieve desirable results by just learning the prompts of $\mathcal{D}$. One step further, adding QO into optimization achieves the best performance. Although other setups introduce much more trainable parameters, they can not surpass the ``LP + QO'' strategy.

\textbf{Effect of Textual Priors.} We study the effect of pre-generated textual priors on LLaMP. We compare three different approaches: without textual priors, using plain responses as the prior, and LLaMP, which takes parsed noun phrases. Tab.~\ref{tab:abl-text-prior} shows that \ours{} can achieve over 81\% on HM without pre-generated priors, while adding parsed noun phrases as textual priors further pushes the HM to 81.27\%.

    \begin{table}[t]
        \centering
        \begin{tabular}{l|ccc|c}
            \toprule
            Method & Priors  & Base & Novel & HM \\
            \midrule
            \multirow{3}{*}{LLaMP}& \xmark &84.90 & 77.59 & 81.08 \\
            & Plain &\textbf{85.26} & 77.56 & 81.22 \\
            & \cellcolor{tabhighlight} NP & \cellcolor{tabhighlight}85.16 & \cellcolor{tabhighlight}\textbf{77.71} &  \cellcolor{tabhighlight}\textbf{81.27} \\
        \bottomrule
        \end{tabular}
        \caption{Ablation Study on Pre-generated Text Priors. \xmark{} refers to ``without textual priors'' and NP stands for noun phrases.}
        \vspace{-1.5mm}
        \label{tab:abl-text-prior}
    \end{table}
    \begin{table}[t]
        \centering
        \begin{tabular}{l|cc|c}
            \toprule
            Method & Base & Novel & HM \\
            \midrule
            LLM Only & 81.74 & 35.82 & 49.81 \\
            \rowcolor{tabhighlight}
            \ours &  \textbf{85.16} &\textbf{77.71} & \textbf{81.27} \\
        \bottomrule
        \end{tabular}
        \caption{The CLIP text encoder helps adaptation.}
        \vspace{-1.5mm}
        \label{tab:abl-bridge}
    \end{table}    
    \begin{table}[t]
        \centering
        \begin{tabular}{l|cc|c}
            \toprule
             Scheme & Base  & Novel & HM \\
             \midrule
             Prompt $\times$ 9 & 84.67 & 77.28 & 80.81\\
             LoRA $\times$ 12 & 84.89 & 77.27 & 80.90 \\
            \midrule
            \rowcolor{tabhighlight}
            Prompt $\times 6$ + LoRA $\times 6$ &  \textbf{85.16} &\textbf{77.71} & \textbf{81.27} \\
        \bottomrule
        \end{tabular}
        \caption{Study on Vision Tuning Scheme. Our hybrid design achieves the best performance.}
        \vspace{-5mm}
        \label{abl:vis-tuning}
    \end{table}
\textbf{CLIP as the bridge.} One may wonder if it is possible to replace CLIP text encoder with a large language model. Here, we study two setups: i) LLM as encoder, which treats the output of the language model, $\bm{\tilde{h}}_l$ as the text feature; ii) LLaMP, which treat $\bm{\tilde{h}}_l$ as part of the text input prompt. Tab.~\ref{tab:abl-bridge} reveals that relying solely on the LLM results in poor accuracy for novel categories. This supports our hypothesis that aligning LLMs with vision encoders generally requires a more extensive dataset. Furthermore, LLaMP's design significantly improve the novel accuracy by 40\%.

\textbf{Vision Training Strategy.} As ViT-16/B has 12 transformer layers, we compare different vision training strategies within \ours{} in Tab.~\ref{abl:vis-tuning}. Apart from the default hybrid scheme, we evaluate setups including prompt learning at first 9 layers (P9), a similar setup to PSRC \cite{khattak2023self}, and LoRA \cite{hu2021lora} in all 12 layers. The results suggest that the scheme leverages the strengths of both prompt learning and LoRA, addressing potential bottlenecks in the vision encoder and enhancing overall performance in \ours.

\begin{figure}
    \centering
    \includegraphics{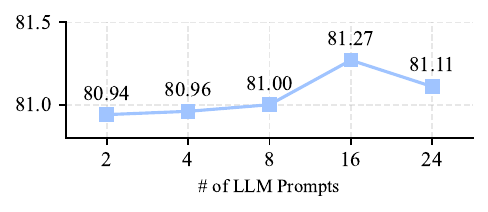}
    \vspace{-8mm}
    \caption{Effect of LLM Prompts on Harmonic Mean. 16 prompts achieve the most balanced performance.}
    \vspace{-3mm}
    \label{fig:llm-prompt}
\end{figure}

\textbf{Number of LLM Prompts.} We vary the number of LLM prompts and study their effects on \ours. As in Fig.~\ref{fig:llm-prompt}, using 16 prompts optimizes the LLM's capabilities, achieving the highest harmonic mean at 81.27\%.

\input{sec/visualization}

\textbf{Visualizations.} In Fig.~\ref{fig:vis}, we visualize the gradient heatmap of input images from FGVCAircraft and EuroSAT, through GradCAM \cite{selvaraju2017grad}. The figure shows that, \ours{} can capture distinctive features that matches LLM's description.

%% file: sec/main_table_compact.tex
\begin{table*}[!t]
\tablestyle{6pt}{0}
\addtolength{\tabcolsep}{-6pt}
    \tabstyle{1.5pt}
    \setlength{\tabcolsep}{6pt}
    \begin{subtable}[t]{\textwidth}
        \centering
        \begin{tabular}{lccc|ccc|ccc|ccc}
            \toprule 
            \multirow{2}[3]{*}{Method} & \multicolumn{3}{c}{\emph{Average}} & \multicolumn{3}{c}{ImageNet \cite{deng2009imagenet}} & \multicolumn{3}{c}{Caltech101 \cite{fei2004learning}} & \multicolumn{3}{c}{OxfordPets \cite{parkhi2012cats}} \\
            \cmidrule(lr){2-4}\cmidrule(lr){5-7}\cmidrule(lr){8-10}\cmidrule(lr){11-13}
            & Base & Novel & \multicolumn{1}{c}{HM} & Base & Novel &\multicolumn{1}{c}{HM} & Base & Novel & \multicolumn{1}{c}{HM} & Base & Novel & \multicolumn{1}{c}{HM}\\
            \midrule
            CLIP \cite{radford2021learning} & 69.34 & 74.22 & 71.70 & 72.43 & 68.14 & 70.22 & 96.84 & {94.00} & 95.40 & 91.17 & 97.26 & 94.12\\
            CoOp \cite{zhou2022learning} & 82.69 & 63.22 & 71.66 & {76.47} & 67.88 & 71.92  & 98.00 & 89.81 & 93.73 & 93.67 & 95.29 & 94.47  \\
            CoCoOp \cite{zhou2022conditional} & 80.47 & 71.69 & 75.83 & 75.98 & {70.43} & {73.10} & 97.96 & 93.81 & {95.84} & 95.20 & 97.69 & {96.43}\\
            KAPT$^*$ \cite{kan2023knowledge} & 78.41 & 70.52 & 74.26 & 71.10 & 65.20 & 68.02  & 97.10 & 93.53 & 95.28 & 93.13 & 96.53 & 94.80 \\ 
            ProDA \cite{lu2022prompt} & 81.56 & 72.30 & 76.65 & 76.66 & 70.54 & 73.47 & 97.74 & 94.36 & 96.02 & 95.43 & 97.76 & 96.58 \\
            MaPLe \cite{khattak2023maple} &	82.28 & 75.14 & 78.55 & 75.40 & 70.32 & 72.72 & 98.27 & 93.23 & 95.68  & 95.43 & \textbf{97.83} & 96.62 \\
            RPO \cite{lee2023read} &	81.13 & 75.00 & 77.78 & 76.60 & \textbf{71.57} & 74.00 & 97.97 & 94.37 & 96.03 & 94.63 & 97.50 & 96.05\\
            PSRC \cite{khattak2023self} & 84.26 & 76.10 & 79.97 & 77.60 & 70.73 & 74.01 & 98.10 & 94.03 & 96.02 & 95.33 & 97.30 & 96.30 \\
            \midrule
            \rowcolor{tabhighlight}
            \ours &  \textbf{85.16} &\textbf{77.71} & \textbf{81.27} &  \textbf{77.99} & 71.27 & \textbf{74.48} & \textbf{98.45} & \textbf{95.85} & \textbf{97.13} & \textbf{96.31} & 97.74 & \textbf{97.02} \\
             $\Delta$ w.r.t. PSRC&  \textcolor{MidnightBlue}{{+0.90}} &  \textcolor{MidnightBlue}{{+1.61}} &  \textcolor{MidnightBlue}{{+1.30}} & \textcolor{MidnightBlue}{+0.39} & \textcolor{MidnightBlue}{+0.54} & \textcolor{MidnightBlue}{+0.47} & \textcolor{MidnightBlue}{+0.35} & \textcolor{MidnightBlue}{+1.82} & \textcolor{MidnightBlue}{+1.11}  & \textcolor{MidnightBlue}{+0.98} & \textcolor{MidnightBlue}{+0.44} & \textcolor{MidnightBlue}{+0.72}\\
            \bottomrule
        \end{tabular}
    \end{subtable}\\
    \begin{subtable}[t]{\textwidth}
        \centering
        \vspace{5pt}
        \begin{tabular}{lccc|ccc|ccc|ccc}
            \toprule 
            \multirow{2}[3]{*}{Method} & \multicolumn{3}{c}{StanfordCars \cite{krause20133d}} & \multicolumn{3}{c}{Flowers102\cite{nilsback2008automated}} & \multicolumn{3}{c}{Food101 \cite{bossard2014food} } & \multicolumn{3}{c}{FGVCAircraft \cite{maji2013fine}} \\
            \cmidrule(lr){2-4}\cmidrule(lr){5-7}\cmidrule(lr){8-10}\cmidrule(lr){11-13}
            & Base & Novel & \multicolumn{1}{c}{HM} & Base & Novel &\multicolumn{1}{c}{HM} & Base & Novel & \multicolumn{1}{c}{HM} & Base & Novel & \multicolumn{1}{c}{HM}\\
            \midrule
            CLIP \cite{radford2021learning} & 63.37 & 74.89 & 68.65 & 72.08 & \textbf{77.80} & 74.83 & 90.10 & 91.22 & 90.66 & 27.19 & 36.29 & 31.09  \\
            CoOp \cite{zhou2022learning} & 78.12 & 60.40 & 68.13 & 97.60 & 59.67 & 74.06 & 88.33 & 82.26 & 85.19 & 40.44 & 22.30 & 28.75 \\
            CoCoOp \cite{zhou2022conditional} & 70.49 & 73.59 & 72.01 & 94.87 & 71.75 & 81.71 & 90.70 & 91.29 & 90.99 & 33.41 & 23.71 & 27.74 \\
            KAPT$^*$ \cite{kan2023knowledge} & 69.47 & 66.20 & 67.79 & 95.00 & 71.20 & 81.40 & 86.13 & 87.06 & 86.59 & 29.67 & 28.73 & 29.19 \\
            ProDA \cite{lu2022prompt} & 72.94 & 74.00 & 73.47 & 95.92 & 72.46 & 82.56 & 90.71 & \textbf{92.05} & 91.38 & 37.44 & 35.61 & 36.50 \\
            MaPLe \cite{khattak2023maple} &	74.70 & 71.20 & 72.91 & 97.70 & 68.68 & 80.66 & 90.30 & 88.57 & 89.43 & 36.90 & 34.13 & 35.46 \\ 
            RPO \cite{lee2023read} & 73.87 & \textbf{75.53} & 74.69 & 94.13 & 76.67 & 84.50 & 90.33 & 90.83 & 90.58 & 37.33 & 34.20 & 35.70 \\
            PSRC \cite{khattak2023self} & 78.27 & 74.97 & 76.58 & \textbf{98.07} & 76.50 & 85.95 & 90.67 & 91.53 & 91.10 & 42.73 & \textbf{37.87} & 40.15 \\
            \midrule
            \rowcolor{tabhighlight}
            \ours &   \textbf{81.56} & 74.54 & \textbf{77.89}  & 97.82 & 77.40 & \textbf{86.42} &  \textbf{91.05} & 91.93 & \textbf{91.49} &  \textbf{47.30} & 37.61 & \textbf{41.90}\\
             $\Delta$ w.r.t. PSRC& \textcolor{MidnightBlue}{+3.29} & \textcolor{Bittersweet}{-0.43} & \textcolor{MidnightBlue}{+1.31} &\textcolor{Bittersweet}{-0.25} & \textcolor{MidnightBlue}{+0.90} & \textcolor{MidnightBlue}{+0.47} & \textcolor{MidnightBlue}{+0.38} & \textcolor{MidnightBlue}{+0.40} & \textcolor{MidnightBlue}{+0.39} & \textcolor{MidnightBlue}{+4.57} & \textcolor{Bittersweet}{-0.26} & \textcolor{MidnightBlue}{+1.75} \\
            \bottomrule
        \end{tabular}
    \end{subtable}\\
   \begin{subtable}[t]{\textwidth}
        \centering
        \vspace{5pt}
        \begin{tabular}{lccc|ccc|ccc|ccc}
            \toprule 
            \multirow{2}[3]{*}{Method} & \multicolumn{3}{c}{SUN397 \cite{xiao2016sun}} & \multicolumn{3}{c}{DTD \cite{cimpoi2014describing}} & \multicolumn{3}{c}{EuroSAT \cite{helber2019eurosat}} & \multicolumn{3}{c}{UCF101 \cite{soomro2012ucf101}}  \\
            \cmidrule(lr){2-4}\cmidrule(lr){5-7}\cmidrule(lr){8-10}\cmidrule(lr){11-13}
            & Base & Novel & \multicolumn{1}{c}{HM} & Base & Novel &\multicolumn{1}{c}{HM} & Base & Novel & \multicolumn{1}{c}{HM} & Base & Novel & \multicolumn{1}{c}{HM}\\
            \midrule
            CLIP \cite{radford2021learning} & 69.36 & 75.35 & 72.23 & 53.24 & 59.90 & 56.37 & 56.48 & 64.05 & 60.03 & 70.53 & 77.50 & 73.85 \\
            CoOp \cite{zhou2022learning} & 80.60 & 65.89 & 72.51 & 79.44 & 41.18 & 54.24 & 92.19 & 54.74 & 68.69 & 84.69 & 56.05 & 67.46 \\
            CoCoOp \cite{zhou2022conditional} & 79.74 & 76.86 & 78.27 & 77.01 & 56.00 & 64.85 & 87.49 & 60.04 & 71.21 & 82.33 & 73.45 & 77.64 \\
            KAPT$^*$ \cite{kan2023knowledge} & 79.40 & 74.33 & 76.78 & 75.97 & 58.30 & 65.97 & 84.80 & 67.57 & 75.21 & 80.83 & 67.10 & 73.33 \\
            ProDA \cite{lu2022prompt} & 80.82 & 78.70 & 79.75 & 80.36 & 59.18 & 68.16 & \textbf{94.07} & 73.23 & 82.35 & 83.00 & 78.66 & 80.77 \\
            MaPLe \cite{khattak2023maple} & 78.47 & 76.93 & 77.79 & 80.67 & 56.48 & 66.44 & 83.90 & 66.00 & 73.88 & 85.23 & 71.97 & 78.04 \\
            RPO \cite{lee2023read} & 80.60 & 77.80 & 79.18 & 76.70 & 62.13 & 68.61 & 86.63 & 68.97 & 76.79 & 83.67 & 75.43 & 79.34 \\
            PSRC \cite{khattak2023self} & 82.67 & 78.47 & 80.52 & 83.37 & 62.97 & 71.75 & 92.90 & 73.90 & 82.32 & 87.10 & 78.80 & 82.74 \\
            \midrule
            \rowcolor{tabhighlight}
            \ours &   \textbf{83.41} & \textbf{79.90} & \textbf{81.62} &  \textbf{83.49} & \textbf{64.49} & \textbf{72.77} &  91.93 & \textbf{83.66} & \textbf{87.60} & \textbf{87.13} & \textbf{80.66} & \textbf{83.77}\\
             $\Delta$ w.r.t. PSRC& \textcolor{MidnightBlue}{+0.74} & \textcolor{MidnightBlue}{+1.43} & \textcolor{MidnightBlue}{+1.10}  & \textcolor{MidnightBlue}{+0.12} & \textcolor{MidnightBlue}{+1.52} & \textcolor{MidnightBlue}{+1.02} & \textcolor{Bittersweet}{-0.97} & \textcolor{MidnightBlue}{+9.76} & \textcolor{MidnightBlue}{+5.28} & \textcolor{MidnightBlue}{+0.03} & \textcolor{MidnightBlue}{+1.86} & \textcolor{MidnightBlue}{+1.03} \\
            \bottomrule
        \end{tabular}
    \end{subtable}
    ~
    \caption{\small\textbf{Comparison with state-of-the-art methods on base-to-novel generalization}. LLaMP shows strong generalization results over previous approaches on 11 image classification tasks. Absolute gains over PSRC are indicated in \textcolor{MidnightBlue}{blue}. $^*$KAPT is trained with ViT-B/32 image encoder instead of ViT-B/16.}
    \label{table:comparision_with_cocoop}
    \vspace{-5mm}
\end{table*}

%% file: sec/visualization.tex
\begin{figure}[t]
\centering
\resizebox{\linewidth}{!}
{
\begin{tabular}{m{2.95cm}<{\centering}cm{2.95cm}<{\centering}}
\toprule
\bf Image & \bf Noun Phrases & \bf Heatmap \\
\midrule
\multicolumn{3}{c}{Classname: \textit{An-12}} \\
\midrule
\includegraphics[width=2.75cm]{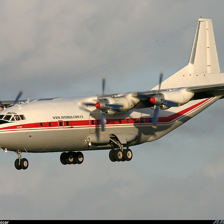}& 
\large\makecell{four engines,\\ turboprop aircraft,\\ large vertical fin} &
\includegraphics[width=2.75cm]{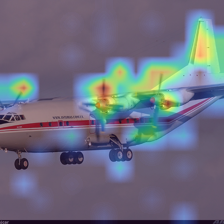} \\
\midrule
\multicolumn{3}{c}{Classname: \textit{Industrial Buildings}} \\
\midrule
\includegraphics[width=2.75cm]{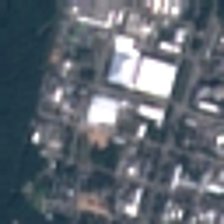}&
\large \makecell{ a cluster, \\ rectangular structures,\\ flat roofs,\\ straight lines} &
\includegraphics[width=2.75cm]{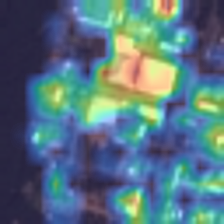} \\
\bottomrule
\end{tabular}}
\caption{Visualization of LLaMP Predictions by GradCAM \cite{selvaraju2017grad}}
\label{fig:vis}
\vspace{-8mm}
\end{figure}

%% file: sec/5_con.tex
\section{Conclusion}

Our study shows that the encyclopedic knowledge from LLMs is beneficial for low-shot image classification as extra class-specific information. To leverage such knowledge, we propose \ours{}, a framework that adapts LLMs as prompt learners for the CLIP model. Over two common low-shot scenarios: zero-shot generalization and few-shot learning, \ours{} demonstrates notable improvements compared with previous state-of-the-arts on a spectrum of datasets.

\textbf{Limitations.} While \ours{} reveals an effective way in leveraging LLMs' knowledge, both modalities, vision and language, only interact at the finest feature level. Given the broader LLM-aided knowledge from the language side, the performance can be potentially further improved by introducing language priors at earlier vision encoding stages.

\section*{Acknowledgment}
This research was supported, in part, by the Office of Naval Research under grant \#N00014-21-1-2802.

%% file: sec/X_suppl.tex
\clearpage
\setcounter{page}{1}
\maketitlesupplementary

\section*{Introduction}

In the supplementary material, we provide extra discussions that did not fit in the main paper due to the space limitation, including i) Ablation study on the textual augmentation described in Sec 3.3; ii) Few-shot classification results on 8/4/2/1 shots with comparisons with previous methods.

\section*{Ablation on Textual Augmentations}
As discussed in Sec 3.3, we perform textual augmentations in two steps: i) When computing $\hat{\bm{g}}$, we replace the original template ``\texttt{A photo of [STH]}'', with ``\texttt{A photo of [STH] with [NP]}'', thereby enriching the descriptive content with noun phrases extracted from LLM responses; ii) We create new LLM prompt templates similar to ``\texttt{In one sentence, describe the distinctive appearance of [STH]}'' through GPT-4 \cite{openai2023gpt4}, and average the scores for final prediction. 

We show ablation study results in Tab.~\ref{tab:abl-ta}, with TA1 and TA2 referring to the step i) and ii) mentioned above. Results show that, even without textual augmentations, LLaMP still outperforms PSRC, the previous state-of-the-art, by 0.68\% on base accuracy, 0.94\% on novel accuracy and 0.83\% on the harmonic mean. 
Moreover, we observe that both augmentation steps further improve the performance of LLaMP. More specifically, TA1 improves the HM by 0.35\% while TA2 brings in another boost of 0.12\%.

\begin{table}[h]
    \centering
    \begin{tabular}{lcc|cc|c}
        \toprule
        Method & TA1 & TA2 & Base & Novel & HM \\
        \midrule
        PSRC \cite{khattak2023self} & & & 84.26 & 76.10 & 79.97 \\
        \midrule 
        \multirow{4}{*}{\ours}  & & & 84.94 & 77.04 & 80.80 \\
        & & \cmark & 84.78 & 77.31 & 80.86 \\
        & \cmark  & & 85.16 & 77.50 & 81.15 \\
        & \cmark & \cmark &\cellcolor{tabhighlight}\textbf{85.16} &\cellcolor{tabhighlight}\textbf{77.71} & \cellcolor{tabhighlight}\textbf{81.27} \\
    \bottomrule
    \end{tabular}
    \caption{Ablation study on textual augmentations.}
    \label{tab:abl-ta}
\end{table}

\section*{Few-shot Classification}
In addition to the 16-shot classification result reported in the main paper, we present few-shot classification results with with 8/4/2/1 numbers of shots in Tab.~\ref{tab:extra-xd} and compare   \ours{}  against previous baseline models. 

Results in Tab.~\ref{tab:extra-xd} show that \ours{} outperforms previous SOTAs under all settings, on average of all 11 benchmarks, with 0.88\% improvement with 8 shots. In particular, we observe that \ours{} surpasses PSRC \cite{khattak2023self} consistently on FGVCAircraft (Aircraft) \cite{maji2013fine} and Food \cite{bossard2014food} with all numbers of shots. Such observation aligns with our argument in the main paper that the knowledge from LLMs provides richer semantic information for fine-grained classification. 

\begin{table*}[t]
    \tabstyle{2.5pt}
    \begin{subtable}[t]{\textwidth}
        \centering
        \begin{tabular}{l c ccccccccccc}
        \toprule
        & \multicolumn{12}{c}{8-Shot Classification} \\ \cmidrule(lr){2-13}
        & \footnotesize\rotatebox{60}{\emph{Average}} & \footnotesize\rotatebox{60}{ImageNet \cite{deng2009imagenet}} & \footnotesize\rotatebox{60}{Caltech \cite{fei2004learning}} & \footnotesize\rotatebox{60}{Pets \cite{parkhi2012cats}} & \footnotesize\rotatebox{60}{Cars \cite{krause20133d}} & \footnotesize\rotatebox{60}{Flowers \cite{nilsback2008automated}} & \footnotesize\rotatebox{60}{Food \cite{bossard2014food}} & \footnotesize\rotatebox{60}{Aircraft \cite{maji2013fine}} & \footnotesize\rotatebox{60}{SUN397 \cite{xiao2016sun}} & \footnotesize\rotatebox{60}{DTD \cite{cimpoi2014describing}} & \footnotesize\rotatebox{60}{EuroSAT \cite{helber2019eurosat}} & \footnotesize\rotatebox{60}{UCF101 \cite{soomro2012ucf101}}  \\
        \midrule
        CLIP \cite{radford2021learning} & 74.47 & 62.23 & 93.41 & 78.36 & 73.67 & 96.10 & 79.79 & 39.35 & 69.08 & 63.46 & 84.43 & 79.34 \\
        CoOp \cite{zhou2022learning} & 76.98 & 70.63 & 94.37 & 91.27 & 79.30 & 94.97 & 82.67 & 39.00 & 71.53 & 64.77 & 78.07 & 80.20 \\
        CoCoOp \cite{zhou2022conditional} & 72.96 & 70.63 & 95.04 & 93.45 & 70.44 & 84.30 & 86.97 & 26.61 & 70.84 & 58.89 & 68.21 & 77.14 \\
        MaPLe \cite{khattak2023maple} & 78.89 & 70.30 & 95.20 & 92.57 & 79.47 & 95.80 & 83.60 & 42.00 & 73.23 & 66.50 & 87.73 & 81.37 \\
        PSRC \cite{khattak2023self} & 80.69 &\bf 72.33 & 95.67 & 93.50 & 80.97 & \bf  96.27 & 86.90 & 43.27 & \bf  75.73 & 69.87 & 88.80 & \bf 84.30 \\
        \midrule
        \rowcolor{tabhighlight}
        \ours & \bf 81.57 & 72.30 & \bf 96.57 & \bf 93.69 & \bf  82.15 & 96.20 & \bf  87.39 & \bf  47.48 & 75.18 & \bf  71.14 & \bf  91.15 & 84.06 \\

        \bottomrule
        \end{tabular}
    \end{subtable}

\begin{subtable}[t]{\textwidth}
        \centering
        \begin{tabular}{l c ccccccccccc}
        \toprule
        & \multicolumn{12}{c}{4-Shot Classification} \\ \cmidrule(lr){2-13}
        & \footnotesize\rotatebox{60}{\emph{Average}} & \footnotesize\rotatebox{60}{ImageNet \cite{deng2009imagenet}} & \footnotesize\rotatebox{60}{Caltech \cite{fei2004learning}} & \footnotesize\rotatebox{60}{Pets \cite{parkhi2012cats}} & \footnotesize\rotatebox{60}{Cars \cite{krause20133d}} & \footnotesize\rotatebox{60}{Flowers \cite{nilsback2008automated}} & \footnotesize\rotatebox{60}{Food \cite{bossard2014food}} & \footnotesize\rotatebox{60}{Aircraft \cite{maji2013fine}} & \footnotesize\rotatebox{60}{SUN397 \cite{xiao2016sun}} & \footnotesize\rotatebox{60}{DTD \cite{cimpoi2014describing}} & \footnotesize\rotatebox{60}{EuroSAT \cite{helber2019eurosat}} & \footnotesize\rotatebox{60}{UCF101 \cite{soomro2012ucf101}}  \\
        \midrule
        CLIP \cite{radford2021learning} & 68.01 & 54.85 & 92.05 & 71.17 & 63.38 & 92.02 & 73.19 & 32.33 & 63.00 & 55.71 & 77.09 & 73.28 \\
        CoOp \cite{zhou2022learning} & 74.02 & 68.73 & 94.40 & 92.57 & 74.47 & 92.17 & 84.47 & 30.83 & 69.97 & 58.70 & 70.80 & 77.10 \\
        CoCoOp \cite{zhou2022conditional} & 71.21 & 70.39 & 94.98 & 92.81 & 69.39 & 78.40 & 86.88 & 24.79 & 70.21 & 55.04 & 65.56 & 74.82 \\
        MaPLe \cite{khattak2023maple} & 75.37 & 67.70 & 94.43 & 91.90 & 75.30 & 92.67 & 81.77 & 34.87 & 70.67 & 61.00 & 84.50 & 78.47 \\
        PSRC \cite{khattak2023self} & 78.35 & 71.07 & 95.27 & 93.43 & \bf 77.13 & 93.87 & 86.17 & 37.47 & 74.00 & 65.53 & \bf 86.30 & 81.57 \\
        \midrule
        \rowcolor{tabhighlight}
        \ours & \bf 78.83 & \bf 71.37 & \bf 95.84 & \bf 93.61 & 76.79 & \bf 93.96 & \bf 87.17 & \bf 40.02 &\bf  74.05 & \bf 66.37 & 86.16 & \bf 81.80 \\
        \bottomrule
        \end{tabular}
\end{subtable}

\begin{subtable}[t]{\textwidth}
    \tabstyle{2.5pt}
        \begin{tabular}{l c ccccccccccc}
        \toprule
        & \multicolumn{12}{c}{2-Shot Classification} \\ \cmidrule(lr){2-13}
        & \footnotesize\rotatebox{60}{\emph{Average}} & \footnotesize\rotatebox{60}{ImageNet \cite{deng2009imagenet}} & \footnotesize\rotatebox{60}{Caltech \cite{fei2004learning}} & \footnotesize\rotatebox{60}{Pets \cite{parkhi2012cats}} & \footnotesize\rotatebox{60}{Cars \cite{krause20133d}} & \footnotesize\rotatebox{60}{Flowers \cite{nilsback2008automated}} & \footnotesize\rotatebox{60}{Food \cite{bossard2014food}} & \footnotesize\rotatebox{60}{Aircraft \cite{maji2013fine}} & \footnotesize\rotatebox{60}{SUN397 \cite{xiao2016sun}} & \footnotesize\rotatebox{60}{DTD \cite{cimpoi2014describing}} & \footnotesize\rotatebox{60}{EuroSAT \cite{helber2019eurosat}} & \footnotesize\rotatebox{60}{UCF101 \cite{soomro2012ucf101}}  \\
        \midrule
        CLIP \cite{radford2021learning} & 57.98 & 44.88 & 89.01 & 58.37 & 50.28 & 85.07 & 61.51 & 26.41 & 53.70 & 40.76 & 61.98 & 65.78 \\
        CoOp \cite{zhou2022learning} & 70.65 & 67.07 & 93.07 & 89.80 & 70.50 & 87.33 & 84.40 & 26.20 & 66.53 & 53.60 & 65.17 & 73.43 \\
        CoCoOp \cite{zhou2022conditional} & 67.65 & 69.78 & 94.82 & 92.64 & 68.37 & 75.79 & 86.22 & 15.06 & 69.03 & 52.17 & 46.74 & 73.51 \\
        MaPLe \cite{khattak2023maple} & 72.58 & 65.10 & 93.97 & 90.87 & 71.60 & 88.93 & 81.47 & 30.90 & 67.10 & 55.50 & 78.30 & 74.60 \\
        PSRC \cite{khattak2023self} & 75.29 & 69.77 & 94.53 & 92.50 & \bf 73.40 & \bf 91.17 & 85.70 & 31.70 & 71.60 & 59.97 & 79.37 & 78.50 \\
        \midrule
        \rowcolor{tabhighlight}
        \ours & \bf 75.89 & \bf 70.12 & \bf 95.66 & \bf 92.75 & 72.20 & 89.16 & \bf 86.33 & \bf 33.41 & \bf 72.64 & \bf 61.29 & \bf 81.71 & \bf 79.56 \\
        \bottomrule
        \end{tabular}
\end{subtable}

\begin{subtable}[t]{\textwidth}
        \centering
    \tabstyle{2.5pt}
        \begin{tabular}{l c ccccccccccc}
        \toprule
        & \multicolumn{12}{c}{1-Shot Classification} \\ \cmidrule(lr){2-13}
        & \footnotesize\rotatebox{60}{\emph{Average}} & \footnotesize\rotatebox{60}{ImageNet \cite{deng2009imagenet}} & \footnotesize\rotatebox{60}{Caltech \cite{fei2004learning}} & \footnotesize\rotatebox{60}{Pets \cite{parkhi2012cats}} & \footnotesize\rotatebox{60}{Cars \cite{krause20133d}} & \footnotesize\rotatebox{60}{Flowers \cite{nilsback2008automated}} & \footnotesize\rotatebox{60}{Food \cite{bossard2014food}} & \footnotesize\rotatebox{60}{Aircraft \cite{maji2013fine}} & \footnotesize\rotatebox{60}{SUN397 \cite{xiao2016sun}} & \footnotesize\rotatebox{60}{DTD \cite{cimpoi2014describing}} & \footnotesize\rotatebox{60}{EuroSAT \cite{helber2019eurosat}} & \footnotesize\rotatebox{60}{UCF101 \cite{soomro2012ucf101}}  \\
        \midrule
        CLIP \cite{radford2021learning} & 45.83 & 32.13 & 79.88 & 44.06 & 35.66 & 69.74 & 43.96 & 19.61 & 41.58 & 34.59 & 49.23 & 53.66 \\
        CoOp \cite{zhou2022learning} & 67.56 & 66.33 & 92.60 & 90.37 & 67.43 & 77.53 & 84.33 & 21.37 & 66.77 & 50.23 & 54.93 & 71.23 \\
        CoCoOp \cite{zhou2022conditional} & 66.79 & 69.43 & 93.83 & 91.27 & 67.22 & 72.08 & 85.65 & 12.68 & 68.33 & 48.54 & 55.33 & 70.30 \\
        MaPLe \cite{khattak2023maple} & 69.27 & 62.67 & 92.57 & 89.10 & 66.60 & 83.30 & 80.50 & 26.73 & 64.77 & 52.13 & 71.80 & 71.83 \\
        PSRC \cite{khattak2023self} & 72.32 & 68.13 & 93.67 & \bf 92.00 & 69.40 & \bf 85.93 & 84.87 & 27.67 & 69.67 & \bf 56.23 & \bf 73.13 & 74.80 \\
        \midrule
        \rowcolor{tabhighlight}
        \ours & \bf 72.42 & \bf 69.12 & \bf 94.59 & 91.91 & \bf 70.02 & 84.03 & \bf 85.83 & \bf 30.39 & \bf 69.69 & 54.98 & 70.36 & \bf 75.72 \\
        \bottomrule
        \end{tabular}
\end{subtable}
\caption{Few shot classification results with 8/4/2/1 shots. All numbers, excepts ours, are obtained from \cite{khattak2023self}.}
\label{tab:extra-xd}
\end{table*}